\documentclass[letterpaper, 10 pt, conference]{./ieeeconf}

\IEEEoverridecommandlockouts                              %

\usepackage[utf8]{inputenc}
\usepackage{graphicx}
\usepackage{url}
\usepackage[linesnumbered]{algorithm2e}
\usepackage{mathtools}
\usepackage{hyperref}
\usepackage{textcomp}
\usepackage{gnuplot-lua-tikz}
\usepackage[style=ieee,autocite=footnote,maxcitenames=2,minnames=2,maxalphanames=4,minalphanames=3,natbib=true,url=false, doi=false]{biblatex}
\usepackage[caption=false, font=footnotesize]{subfig}

\renewcommand*{\bibfont}{\small}
\DefineBibliographyStrings{english}{%
    andothers = {et\addabbrvspace al\adddot}
}
\begin{document}

\title{SLAM auto-complete: completing a robot map using an emergency map}

\author{Malcolm Mielle, Martin Magnusson, Henrik Andreasson, Achim J\@. 
Lilienthal
   \thanks{Center of Applied Autonomous Sensor Systems (AASS), 
{\"O}rebro University, Sweden.
     {\tt firstname.lastname@oru.se}}%
   \thanks{This work was funded in part by
     the EU H2020 project
     SmokeBot (ICT-23-2014 645101) and by the Swedish Knowledge Foundation under contract number 20140220 (AIR)
}}

\maketitle

\begin{abstract}

In search and rescue missions, time is an important factor; fast navigation and quickly acquiring situation awareness might be matters of life and death.
Hence, the use of robots in such scenarios has been restricted by the time needed to explore and build a map. One way to speed up exploration and mapping is to reason about unknown parts of the environment using prior information.
While previous research on using external priors for robot mapping mainly focused on accurate maps or aerial images, such data are not always possible to get, especially indoor.
We focus on emergency maps as priors for robot mapping since they are easy to get and already extensively used by firemen in rescue missions. 
However, those maps can be outdated, information might be missing, and the scales of rooms are typically not consistent. 
We have developed a formulation of graph-based SLAM that incorporates information from an emergency map.
The graph-SLAM is optimized using a combination of robust kernels, fusing the emergency map and the robot map into one map, even when faced with scale inaccuracies and inexact start poses.

We typically have more than 50\% of wrong correspondences in the settings studied in this paper, and the method we propose correctly handles them. Experiments in an office environment show that we can handle up to 70\% of wrong correspondences and still get the expected result. 
The robot can navigate and explore while taking into account places it has not yet seen. We demonstrate this in a test scenario and also show that the emergency map is enhanced by adding information not represented such as closed doors or new walls.

\end{abstract}

\IEEEpeerreviewmaketitle

\section{Introduction}

In a search and rescue scenario, speed and efficiency are key. First responders need to locate victims, and get them out (alive) of the disaster site as quickly as they can. Each of those tasks usually takes up several hours, in harsh and dangerous conditions, putting the lives of first responders at risk. Autonomous robots could reduce the time spent by first responders on the disaster site, and speed-up operations. On the other hand, one does not want to spend much time in exploration and map building, two critical tasks needed by robots.
One way to tackle this problem is to integrate prior information into Simultaneous Localization And Mapping (SLAM), to reason on still unknown parts of the environment

For indoor environments, emergency maps are probably the easiest prior maps to get. 
A use case is for firemen, who easily have access to emergency maps during their missions. 
However, the maps can be outdated, new changes to the building will not be represented, and the scale might not be uniform, to make the map easier to interpret.

Previous works on SLAM with prior information focused on using either a topological map depicting objects \cite{shah_qualitative_2013} or a map representing an environment with no distortions or errors, e.g. aerial maps \cite{persson_fusion_2008, parsley_towards_2010,  kummerle_large_2011} or highly accurate models of the environment~\cite{parsley_exploiting_2011, vysotska_exploiting_2016}. 
While the methods mentioned above can deal with outdated data, they assume that the prior is a metrically accurate map. It is not straight forward to integrate emergency maps in SLAM using approaches from the literature~\cite{shah_qualitative_2013, persson_fusion_2008, kummerle_large_2011, parsley_towards_2010, parsley_exploiting_2011, vysotska_exploiting_2016} due to the non-uniform scale. 

We aim at enhancing the map built during SLAM (i.e. SLAM map) by using information from
an emergency map: 
the SLAM map is completed with information from the emergency map, and the emergency map inaccuracies are corrected using the sensors' measurements. 

\section{Method outline and contributions}

\begin{figure}[t]
    \begin{center}
        \includegraphics[width = 7cm]{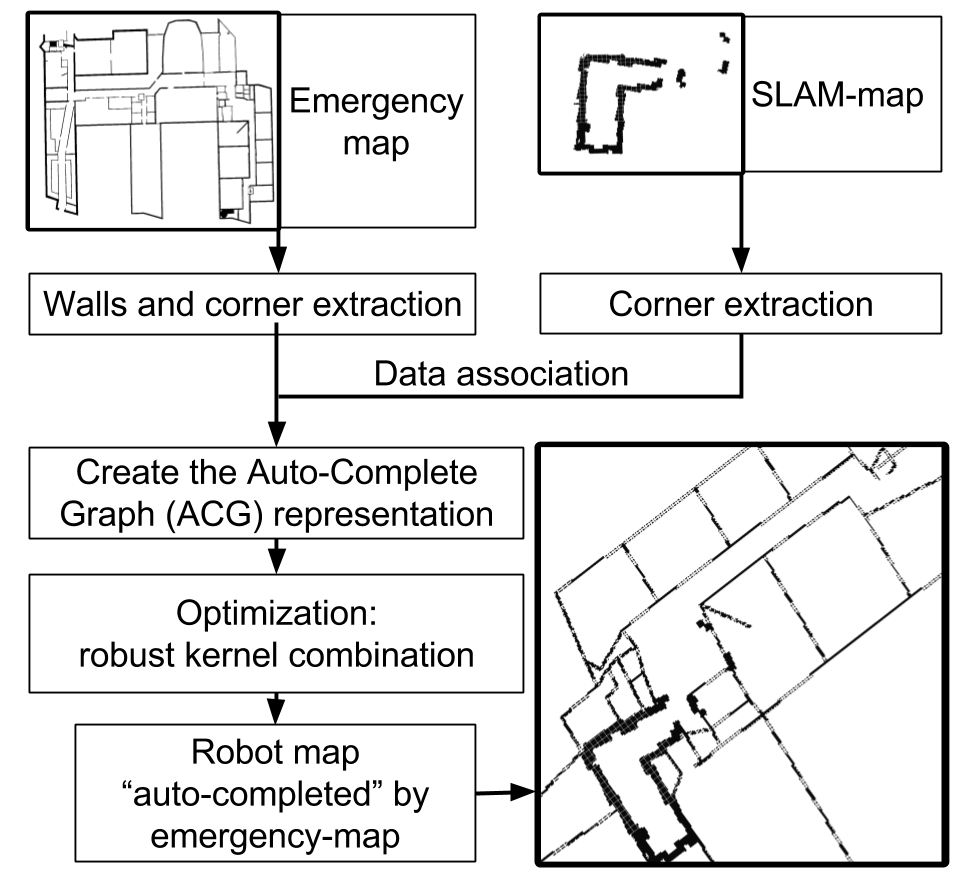}
        \caption{Auto-complete process.}
        \label{fig:process}
    \end{center}
    \vspace{-8mm}
\end{figure}

We create a graph representation, that we call the auto-complete graph (ACG), fusing information from the SLAM map and the emergency map into one. 
Nodes in the ACG are the corners from the emergency map, the robot poses, and the corners in the SLAM map. Graph edges are added between connected corners in the emergency map (along the walls), between consecutive robot poses (from scan registration), between corners in the SLAM map and robot poses (observations edges), and between corresponding corners in both maps. The ACG is optimized, fusing information from the emergency and SLAM maps. The whole process is illustrated in Fig~\ref{fig:process}.

The contributions of the paper are :

\begin{itemize}
    \item A formulation of graph-based SLAM that incorporates information from a rough prior map with uncertainties in scale and detail level.
    \item An optimization strategy adapted to the new graph formulation, based on a combination of robust kernels.
\end{itemize}

We also demonstrate a method for extracting corners in maps using the NDT (normal-distributions transform) representation.

Related works are covered in more detail in Section~\ref{sec:prev}. We show how to build the ACG in Section~\ref{sec:method} and how to optimize it in Section~\ref{sec:optimization}. The method is tested and validated in Section~\ref{sec:parameval} and Section~\ref{sec:exp}.

\section{Related work}
\label{sec:prev}

\textcite{vysotska_exploiting_2016} match maps obtained from OpenStreetMap onto the robot scans. They use the robot position and ICP~\cite{besl_method_1992} to introduce a correcting factor in the error function used in the graph-SLAM\@. The method allows up to $\pm 40^\circ$ of error in orientation between the map and the robot's heading for outdoor maps but can not correct an emergency map since its local scale may not correspond to the robot scans and ICP can not deform the map.

\textcite{parsley_exploiting_2011} integrate planes extracted from Ordnance Survey MasterMap\footnote{\url{https://www.ordnancesurvey.co.uk/business-and-government/products/mastermap-products.html}} as constraints in a graph representation, use gating to remove planes that do not correspond to any planes in the scans, and match corresponding prior and SLAM planes using RANSAC.
While the framework produces a more accurate map than SLAM without prior information, the gating is done using a distance metric, assuming uniform scale. Hence, correspondences between planes from the prior and the SLAM map are assumed to be correct. We assume that correspondences between the emergency and SLAM maps will introduce errors in the graph.
Thus, we constrain the SLAM map using the prior while correcting the prior to complete the SLAM map.

Another idea is to consider the prior map as a topological representation of the environment. \textcite{shah_qualitative_2013} use a human-provided map representing buildings to create waypoints using the Voronoi--Delaunay graph. The prior map is matched onto the robot map to estimate an affine transformation. Indoor, one could use objects, but annotated maps are harder to get than emergency maps. 

\textcite{oswald_speeding-up_2016} developed an exploration strategy that uses hand-made topo-metrical maps
and the traveling salesman problem to find the most efficient global exploration path in the prior.
Since they focus on the exploration method, there is no clear explanation on how to match the prior onto the SLAM map.

Skubic, et al. \cite{skubic_using_2007, skubic_sketch_2003, parekh_scene_2007} use a sketch map interface for navigation.
Objects are closed polygons drawn by the user and are described and matched to objects seen by the robot using histogram of forces. 
Their method can not be used on emergency maps with no objects represented. 

In our previous work~\cite{mielle_using_2016}, we developed a method to find correspondences between a sketch map and a ground truth map. To interpret the sketch map, we use a Voronoi diagram, with a thinning parameter, that we extract as a graph and we use an efficient error-tolerant graph matching algorithm to find correspondences. 
However, the method cannot be used directly on a SLAM map due to their high level of noise.

\textcite{freksa_schematic_2000} use a schematic map to navigate a robot. 
They find correspondences between the schematic map and the environment by matching corners and tested their method on a simulated environment with three rooms.
By definition, their prior map and data association are perfect, making the method too simple for emergency maps.

\textcite{kummerle_large_2011} used aerial maps as prior in a graph-based SLAM\@. They use edges in both modalities and Monte Carlo localization to find correspondences between stereo and three-dimensional range data, and the aerial images. 
However, they assume a constant scale in the aerial image, which we can not do with emergency maps.

\textcite{boniardi_robot_2015} present an approach for robot localization and navigation based on a hand-drawn sketch of the environment. They use an extension of the Monte Carlo localization algorithm to track the robot pose and approximate the deformation between the sketch map and the real-world using two scale factors. 
However, two scale factors will not correct local deformations of the emergency map.

\section{Auto-complete-graph construction}
\label{sec:method}

We developed a graph-based SLAM representation that incorporates information from an emergency map and SLAM map into one map, by using corners as common landmarks. Corners are easy to extract in emergency maps since walls are drawn clearly, making corners salient. 
We describe the method to extract elements from SLAM maps in Section~\ref{subsec:slam_element}, and from emergency maps in Section~\ref{subsec:prior_element}, before presenting the ACG formulation in Section~\ref{subsec:construction}.

\subsection{Processing the SLAM map}
\label{subsec:slam_element}

\begin{figure}[t]
\centering
\begin{tabular}{cc} 
    \subfloat[Full NDT map with corners (in green) detected using a neighborhood size of two.]{\includegraphics[width = 0.4\columnwidth]{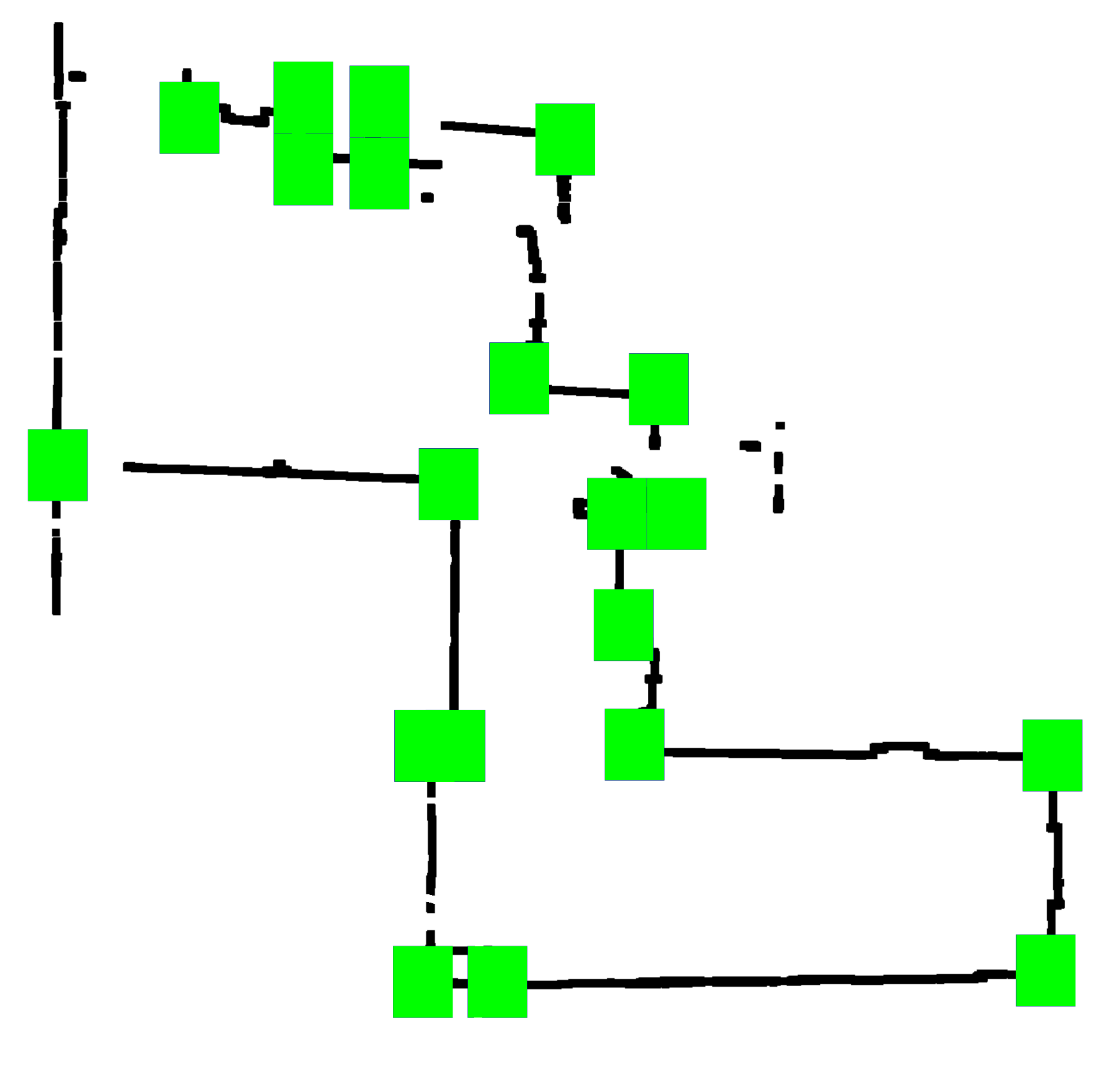}\label{fig:ndtcorners}} &
    \subfloat[When using a neighborhood search size of 1, a NDT-corner was undetected since it is next to an empty cell.]{\includegraphics[width = 0.4\columnwidth]{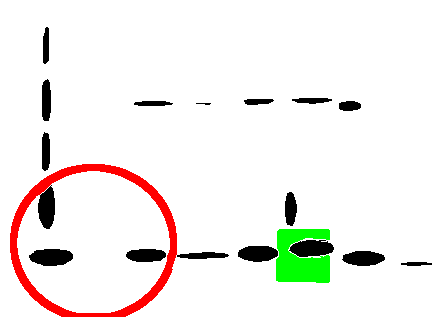}\label{fig:ndtcornersbad}}
    \end{tabular}
    \caption{A NDT map, with the Gaussians in black, and extracted corners, represented by green squares.}
    \label{fig:ndtmaps}
    \vspace{-6mm}
\end{figure}

We use NDT~\cite{biber_normal_2003, stoyanov_normal_2013} as the representation of the robot's map. NDT makes it easy to extract salient corners, and
allows for efficient scan registration~\cite{magnusson_scan_2007, magnusson_beyond_2015}, planning \cite{stoyanov_path_2010} and localization
in both 2D and 3D\@. 
NDT is a grid-based representation where each cell stores a Gaussian distribution representing the shape of the local surface. While mapping an environment with a range scanner, an NDT map can be built incrementally by representing each scan as an NDT grid, then registering and fusing it with the previous map~\cite{stoyanov_normal_2013}. From this, to build the pose-graph representation used later to build the ACG, a partial NDT map is built iteratively until the robot goes further than a certain distance, at which point a new NDT map is started. Each pose where we started to build a partial NDT map is a pose-node in the graph, and possesses its corresponding partial NDT map as an attribute. More on this process is detailed in Section~\ref{subsubsec:elem}.
Note that the approach presented in this paper does not hinge on using NDT as the representation of the robot map. Other representations could be used, as long as it is possible to extract salient corners.
To find corners in each NDT map, we analyze every cell that is occupied, i.e. each cell that has a Gaussian. 
A cell is a corner if:
    1) it has more than one neighbor with a Gaussian,
    2) the main eigenvector of two of those neighbors' Gaussian's form an angle between 80 and 100 degrees.
By calculating the rays' collision point, the estimated position of the corner is determined.
However, this method depends on the size of the neighborhood observed at every point. Some corners can be overlooked if one considers a small neighborhood that does not have enough measurements, as seen in Fig~\ref{fig:ndtcornersbad}. In our work, the neighborhood size considered is 2.
A resulting NDT map with detected corners can be seen in Fig~\ref{fig:ndtcorners}.

\subsection{Processing the emergency map}
\label{subsec:prior_element}

\begin{figure}[t]
\centering
\begin{tabular}{ccc}
        \subfloat[Partial view of an emergency map from a larger picture, taken by a 20MP phone camera.]{\includegraphics[height = 2.6cm]{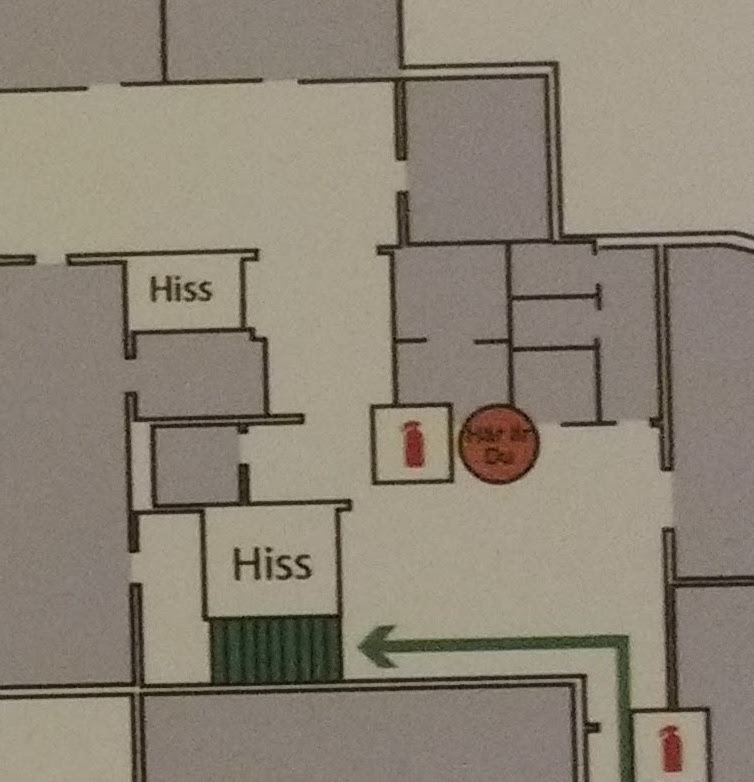}\label{fig:realmap}} &
        \subfloat[Emergency map with only the walls represented. There are some deformations due to icons removal.]{\includegraphics[height = 2.6cm]{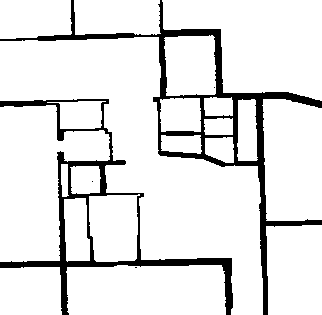}\label{fig:graphextract1}} &
        \subfloat[Graph extracted from the prior.]{\includegraphics[height = 2.6cm]{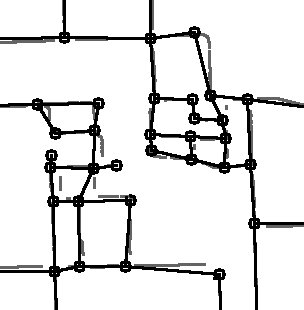}\label{fig:graphextract}}
\end{tabular}
\caption{Pre (Fig~\ref{fig:realmap}) and post processed emergency map (Fig~\ref{fig:graphextract1} and Fig~\ref{fig:graphextract}). Because of the scale uncertainties and artifacts due to symbols, the graph extracted is not an accurate representation of the environment.}
\vspace{-6mm}
\end{figure}
\label{fig:problemexample}

Emergency maps possess few features, as in Fig~\ref{fig:realmap}. The maps usually represent walls and some elements, such as the position of extinguishers, stairs, or toilets. We removed manually those extra elements as they might not be consistently represented between different emergency maps~\cite{dymon_analysis_2003}. 
A corner in the emergency map is, either a place where the line orientation abruptly deviates with an angle of 45\textdegree\,or more, or a place where a line splits into multiple lines, i.e a crossing. Using a line follower algorithm, all corners and the lines between them are extracted from the emergency map,
as in Fig~\ref{fig:graphextract}.
The line follower is able to take into account uneven line thickness of the walls in the thresholded image of the emergency map.

\subsection{Graph formulation}
\label{subsec:construction}

\begin{figure}[t]
    \hspace*{-1.5in}
    \begin{center}
        \includegraphics[width = \columnwidth]{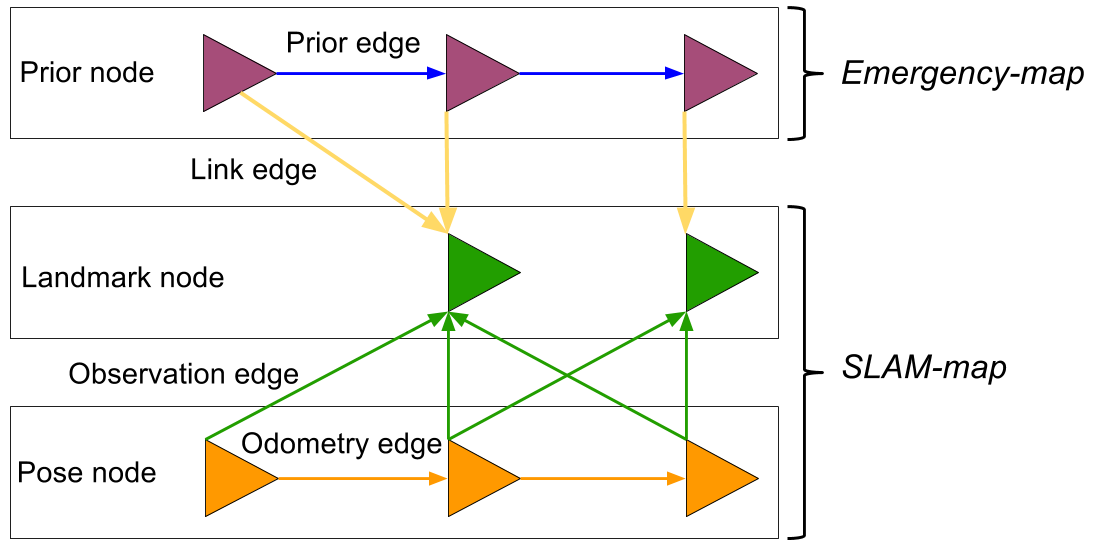}
        \caption{The ACG structure fusing information from the SLAM and emergency maps. Prior-nodes are in purple, prior-edges in blue, pose-nodes and odometry-edge in orange, landmarks and their observations in green, and link-edges in yellow. We will use this color code in all pictures.}
        \label{fig:graphstruct}
    \end{center}
    \vspace{-8mm}
\end{figure}

The ACG is built using the information extracted previously from the SLAM and emergency maps.
The structure of the ACG can be seen in Fig~\ref{fig:graphstruct}.

\subsubsection{Elements from the NDT maps}
\label{subsubsec:elem}
robot poses are added as pose-nodes in the ACG, while odometry-edges are the registrations between partial NDT maps associated to each consecutive pose-node.
Corners extracted from the NDT maps are landmark-nodes in the ACG. The observation measurements between landmarks and the pose-nodes from where they were detected are observation-edges.
Since the partial NDT maps associated with each pose node are rigid, observation-edges must reflect this.
Hence, their covariances' standard deviation is $\sqrt{0.05}$\,m for the $\vec{x}$ and $\vec{y}$ axis, to only allow for small movements of landmark-nodes around their respective pose-node.

\subsubsection{Elements from the emergency map}
\label{subsubsec:elemerg}

the corners from the emergency map are prior-nodes in the graph and walls between them are prior-edges.
We initialize the prior-nodes' positions using a transformation given by two known equivalent points in both maps.
Prior-edges' lengths are computed from the distance between prior-nodes.

While emergency maps have uncertainties in scale and proportion, they are structurally consistent and we aim at preserving that consistency where needed: prior-edges (i.e walls in the emergency map) should be hard to rotate but easy to stretch or shrink. 
Thus, each prior-edge has a covariance with a high value along its main axis but a small one on the perpendicular axis. 
The covariance matrix can be defined with $\Sigma V = V L$ where $\Sigma$ is the covariance matrix, $V$ is the matrix whose columns are the eigenvectors of $\Sigma$ and $L$ is the diagonal matrix whose non-zero elements are the corresponding eigenvalues. By computing $\Sigma = V L V^{-1}$ we obtain the desired translation-covariance.
To align the covariance with the edge axis, we define the first eigenvector as the direction of the edge, while the second eigenvector is a perpendicular vector. We associate a high eigenvalue to the first eigenvector
and a small eigenvalue to the second one. The calculation of the first eigenvalue depends on the original, non-optimized, length of the prior-edge;
we experimented with different values in Section~\ref{subsec:priorcov}. The second eigenvalue is manually set to a small value: 0.005.

\subsubsection{Matching the corners}

to associate each corner extracted from the NDT maps with potential correspondences in the prior map, an edge between every landmark-node and every prior-node is added to the ACG, if the distance between them is less than a certain threshold. Those link-edges represent a zero transformation and their covariance's standard deviation 
is set to $\sqrt{0.5}$\,m on the $\vec{x}$ and $\vec{y}$ axis to account for possible noise. Effectively, the linked corners are allowed to move uniformly around each other, but they can't go far from each other without increasing cost.

\section{Optimization and SLAM back-end}
\label{sec:optimization}

\begin{figure}[t]
\centering
\begin{tabular}{cc}
        \subfloat[Optimization result when using a Huber kernel first followed by DCS. The emergency map is fitted onto the SLAM map.]{\includegraphics[height = 3.5cm]{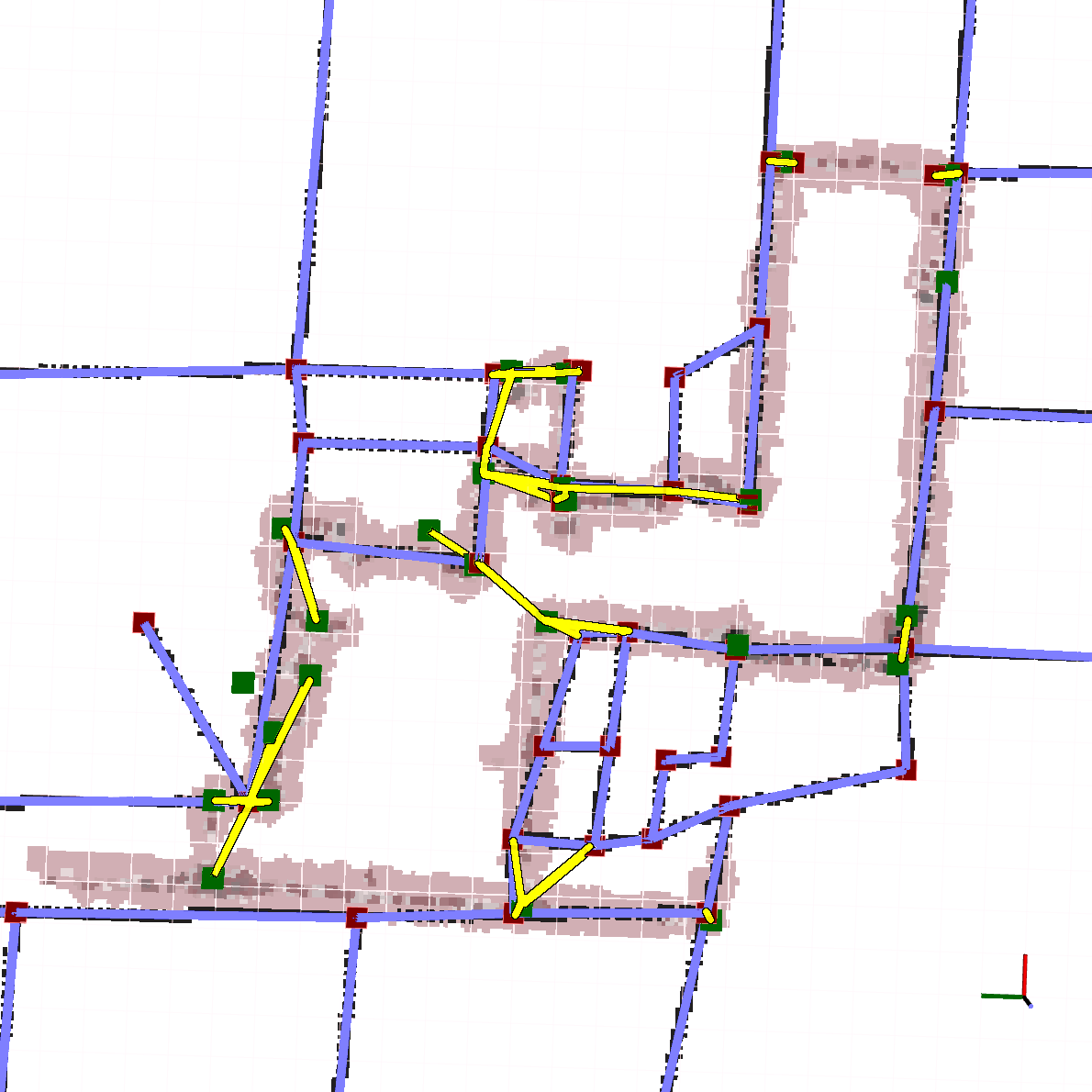}\label{fig:Huberdcs}}&
        \subfloat[Result of the optimization when no robust kernel is used. Large deformations due to incorrect corner association are visible in the red ellipses.]{\includegraphics[height = 3.5cm]{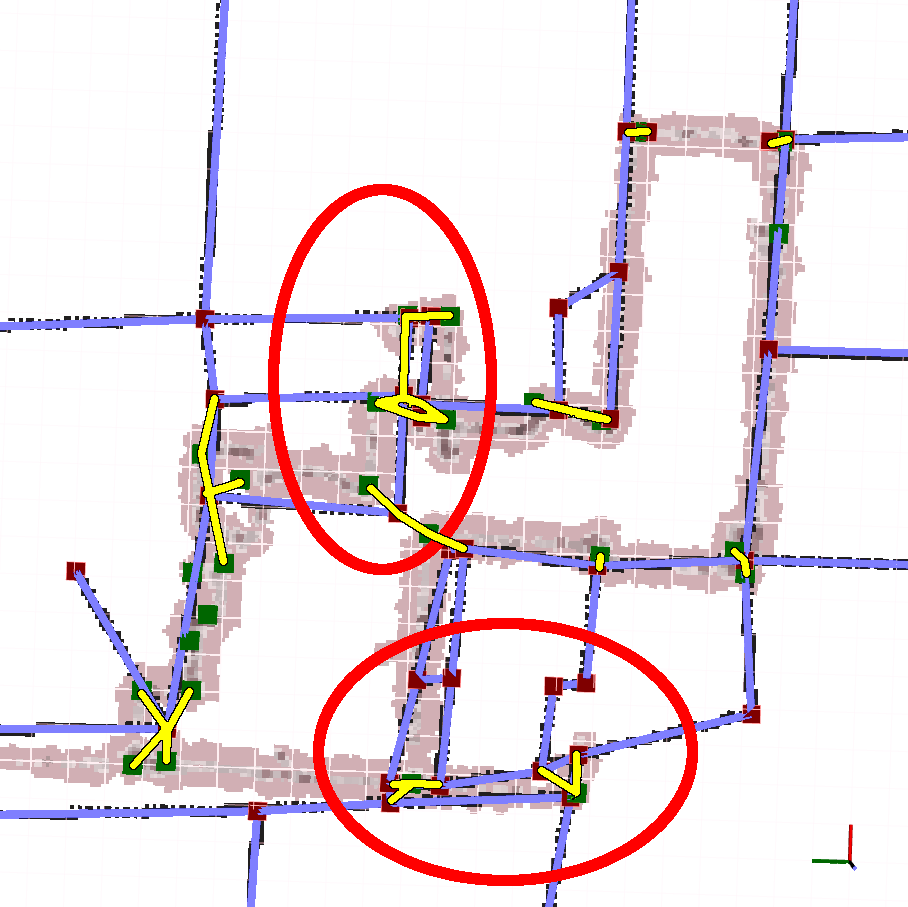}\label{fig:norobustkernel}} \\
        \subfloat[Optimization result when only a Huber kernel is used. Deformations due to incorrect corner association are still visible in the red ellipses.]{\includegraphics[height = 3.5cm]{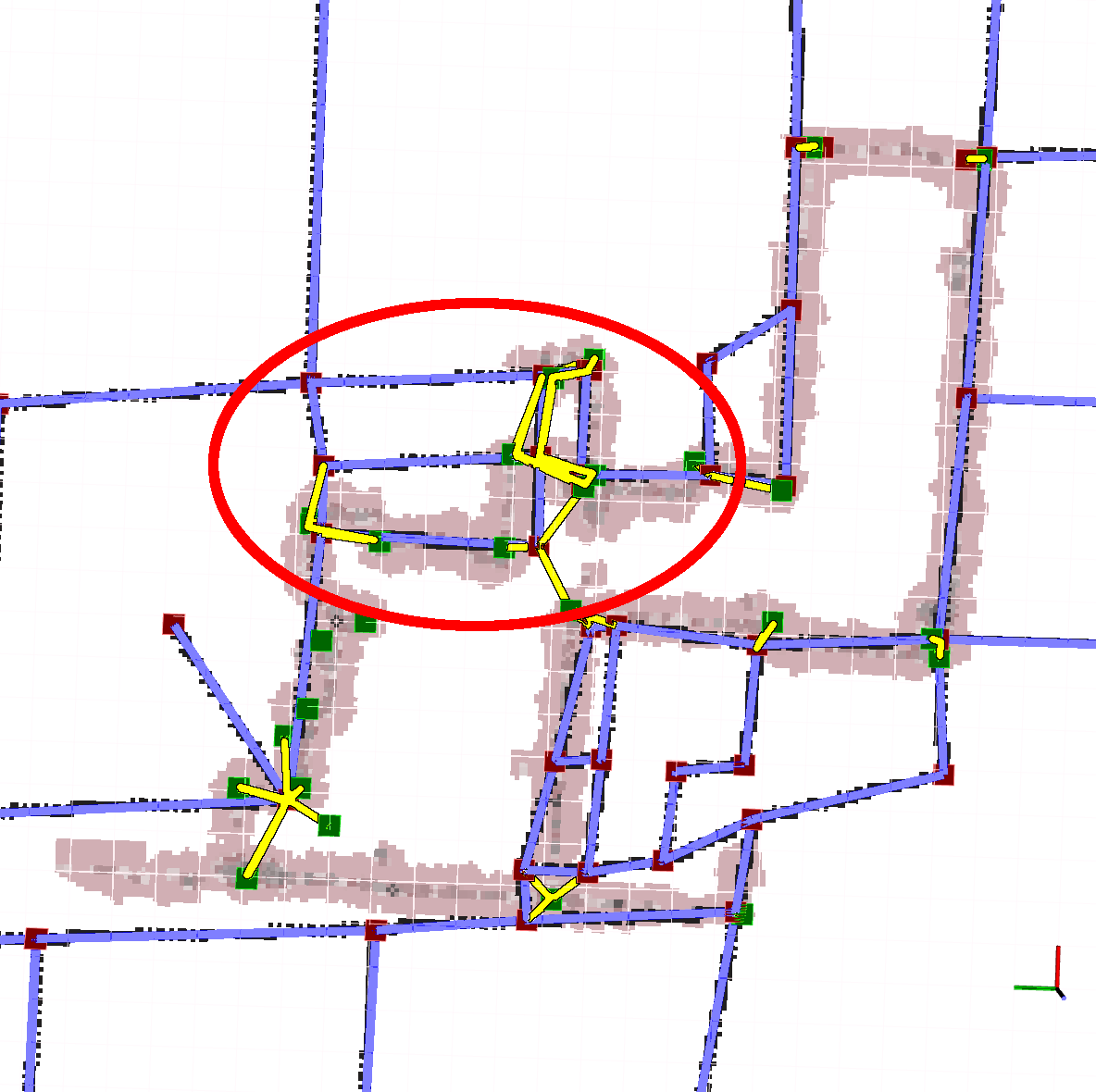}\label{fig:Huber}} &
        \subfloat[Optimization result when only DCS is used. The partial NDT maps have been moved and don't fit together.]{\includegraphics[height = 3.5cm]{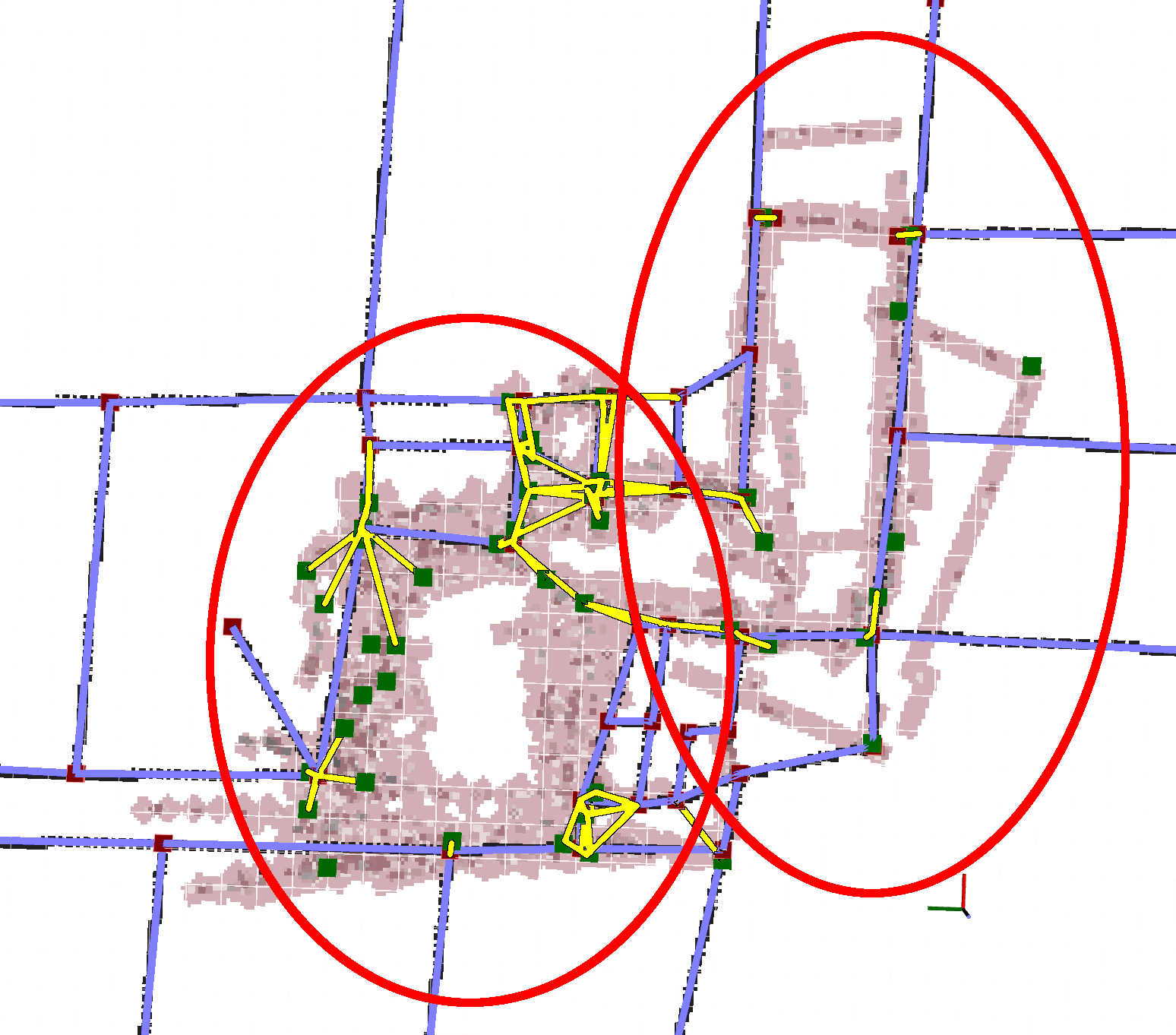}\label{fig:dcs}} 
         
\end{tabular}
\caption{Different results of the optimization using different optimization strategies. Link-edges are in yellow, prior-edges in blue and the SLAM map is in light brown.}
\label{fig:kerneltest}
\vspace{-6.5mm}
\end{figure}

Now that we have a graph representation of the environment, the next step is to optimize it to complete the SLAM map with not yet seen parts of the environment. 
With each new pose-node added to the ACG, we
find all possible correspondences between prior and landmark-nodes, before running 10 optimization iterations with a Huber kernel, followed by 20 iterations with Dynamic Covariance Scaling~\cite{agarwal_robust_2013, agarwal_robust_2015}~(DCS) to increase robustness to the very high number of outlier edges. The Huber kernel is a parabola in the vicinity of zero and increases linearly at a
given level $j * j > k$. Its cost-function is as follow:
\begin{equation}  \label{eq:Hubercost}
    \rho(x) = 
    \begin{cases}
      \frac{x^2}{2}, & \text{if}\ |x|\leq k \\
      k(|x| - \frac{k}{2}), & \text{otherwise}
    \end{cases}
\end{equation} 
While this kernel guarantees unicity of the solution since it is a convergent $\rho$-function, the result is still influenced by incorrect link-edges between non-corresponding corners (Fig~\ref{fig:Huber}). To remove the effect of remaining incorrect link-edges, we use DCS to dynamically scale down the information matrices in edges that introduce a large error in the graph.

\section{Parameter evaluation}
\label{sec:parameval}

We evaluated our method on Örebro University’s dataset~\footnote{\url{http://wiki.ros.org/perception\_oru/Tutorials/Using\%20NDT\%20Fuser\%20to\%20create\%20an\%20NDT\%20map}}, using a picture of the emergency map, taken by a phone, as the prior (see partial image in Fig~\ref{fig:realmap}). A public implementation is available online~\footnote{\url{https://github.com/MalcolmMielle/Auto-Complete-Graph}}.

\subsection{Influence of the robust kernel on the optimization}

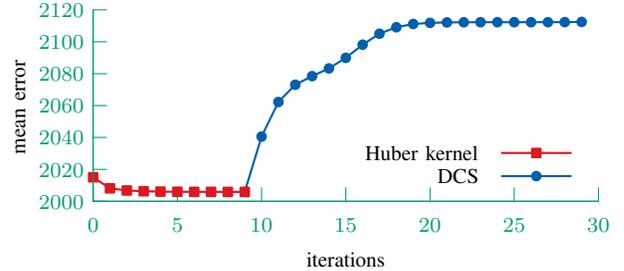
\begin{figure}[t]
\vspace{+2mm}
    \begin{center} 
      \begin{tikzpicture}[gnuplot]
\tikzset{every node/.append style={font={\footnotesize}}}
\gpcolor{rgb color={0.000,0.620,0.451}}
\gpsetlinetype{gp lt border}
\gpsetdashtype{gp dt solid}
\gpsetlinewidth{1.00}
\draw[gp path] (1.228,0.985)--(1.408,0.985);
\node[gp node right] at (1.228,0.985) {$2000$};
\draw[gp path] (1.228,1.409)--(1.408,1.409);
\node[gp node right] at (1.228,1.409) {$2020$};
\draw[gp path] (1.228,1.834)--(1.408,1.834);
\node[gp node right] at (1.228,1.834) {$2040$};
\draw[gp path] (1.228,2.258)--(1.408,2.258);
\node[gp node right] at (1.228,2.258) {$2060$};
\draw[gp path] (1.228,2.682)--(1.408,2.682);
\node[gp node right] at (1.228,2.682) {$2080$};
\draw[gp path] (1.228,3.107)--(1.408,3.107);
\node[gp node right] at (1.228,3.107) {$2100$};
\draw[gp path] (1.228,3.531)--(1.408,3.531);
\node[gp node right] at (1.228,3.531) {$2120$};
\draw[gp path] (1.228,0.985)--(1.228,1.165);
\node[gp node center] at (1.228,0.677) {$0$};
\draw[gp path] (2.348,0.985)--(2.348,1.165);
\node[gp node center] at (2.348,0.677) {$5$};
\draw[gp path] (3.467,0.985)--(3.467,1.165);
\node[gp node center] at (3.467,0.677) {$10$};
\draw[gp path] (4.587,0.985)--(4.587,1.165);
\node[gp node center] at (4.587,0.677) {$15$};
\draw[gp path] (5.707,0.985)--(5.707,1.165);
\node[gp node center] at (5.707,0.677) {$20$};
\draw[gp path] (6.826,0.985)--(6.826,1.165);
\node[gp node center] at (6.826,0.677) {$25$};
\draw[gp path] (7.946,0.985)--(7.946,1.165);
\node[gp node center] at (7.946,0.677) {$30$};
\draw[gp path] (1.228,3.531)--(1.228,0.985)--(7.946,0.985);
\gpcolor{color=gp lt color border}
\node[gp node center,rotate=-270] at (0.246,2.258) {mean error};
\node[gp node center] at (4.587,0.215) {iterations};
\node[gp node right] at (6.478,1.319) {DCS};
\gpcolor{rgb color={0.000,0.376,0.678}}
\gpsetlinewidth{2.00}
\draw[gp path] (6.662,1.319)--(7.578,1.319);
\draw[gp path] (3.243,1.109)--(3.467,1.845)--(3.691,2.305)--(3.915,2.535)--(4.139,2.649)%
  --(4.363,2.752)--(4.587,2.893)--(4.811,3.068)--(5.035,3.214)--(5.259,3.300)--(5.483,3.343)%
  --(5.707,3.358)--(5.931,3.364)--(6.155,3.366)--(6.378,3.367)--(6.602,3.367)--(6.826,3.367)%
  --(7.050,3.367)--(7.274,3.367)--(7.498,3.368)--(7.722,3.371);
\gpsetpointsize{4.00}
\gppoint{gp mark 7}{(3.243,1.109)}
\gppoint{gp mark 7}{(3.467,1.845)}
\gppoint{gp mark 7}{(3.691,2.305)}
\gppoint{gp mark 7}{(3.915,2.535)}
\gppoint{gp mark 7}{(4.139,2.649)}
\gppoint{gp mark 7}{(4.363,2.752)}
\gppoint{gp mark 7}{(4.587,2.893)}
\gppoint{gp mark 7}{(4.811,3.068)}
\gppoint{gp mark 7}{(5.035,3.214)}
\gppoint{gp mark 7}{(5.259,3.300)}
\gppoint{gp mark 7}{(5.483,3.343)}
\gppoint{gp mark 7}{(5.707,3.358)}
\gppoint{gp mark 7}{(5.931,3.364)}
\gppoint{gp mark 7}{(6.155,3.366)}
\gppoint{gp mark 7}{(6.378,3.367)}
\gppoint{gp mark 7}{(6.602,3.367)}
\gppoint{gp mark 7}{(6.826,3.367)}
\gppoint{gp mark 7}{(7.050,3.367)}
\gppoint{gp mark 7}{(7.274,3.367)}
\gppoint{gp mark 7}{(7.498,3.368)}
\gppoint{gp mark 7}{(7.722,3.371)}
\gppoint{gp mark 7}{(7.120,1.319)}
\gpcolor{color=gp lt color border}
\node[gp node right] at (6.478,1.627) {Huber kernel};
\gpcolor{rgb color={0.867,0.094,0.122}}
\draw[gp path] (6.662,1.627)--(7.578,1.627);
\draw[gp path] (1.228,1.304)--(1.452,1.157)--(1.676,1.128)--(1.900,1.118)--(2.124,1.113)%
  --(2.348,1.111)--(2.572,1.110)--(2.796,1.110)--(3.019,1.109)--(3.243,1.109);
\gppoint{gp mark 5}{(1.228,1.304)}
\gppoint{gp mark 5}{(1.452,1.157)}
\gppoint{gp mark 5}{(1.676,1.128)}
\gppoint{gp mark 5}{(1.900,1.118)}
\gppoint{gp mark 5}{(2.124,1.113)}
\gppoint{gp mark 5}{(2.348,1.111)}
\gppoint{gp mark 5}{(2.572,1.110)}
\gppoint{gp mark 5}{(2.796,1.110)}
\gppoint{gp mark 5}{(3.019,1.109)}
\gppoint{gp mark 5}{(3.243,1.109)}
\gppoint{gp mark 5}{(7.120,1.627)}
\gpcolor{rgb color={0.000,0.620,0.451}}
\gpsetlinewidth{1.00}
\draw[gp path] (1.228,3.531)--(1.228,0.985)--(7.946,0.985);
\gpdefrectangularnode{gp plot 1}{\pgfpoint{1.228cm}{0.985cm}}{\pgfpoint{7.946cm}{3.531cm}}
\end{tikzpicture}
      \vspace{-2mm}
      \caption{Convergence plot for our optimization strategy: the mean error is calculated over 25 runs. For each run, optimization with the Huber kernel reaches stability at around 5 iterations, while DCS does after 13 additional iterations. DCS allows some link edges to move further from their mean value so the total error plotted increases after optimization}
        \label{fig:chi}
    \end{center}
    \vspace{-8mm}
\end{figure}

We evaluate the influence of the robust kernels and confirm that using a combination of a Huber Kernel and DCS is a good optimization strategy. 

In Fig~\ref{fig:Huberdcs}, one can see the result of the optimization when using a Huber kernel for 10 iterations first, followed by 20 iterations with DCS: the emergency map is fitted onto the SLAM map and we obtain the expected result. 
Fig~\ref{fig:chi} shows the mean error value at each iteration step, over 28 runs. One can see that the Huber kernel reaches stability in around 5 iterations and DCS reaches stability in around 13 iterations. 

With no robust kernel, the emergency map is not correctly fitted on the SLAM map: rooms from the prior are deformed in a non-realistic way, as can be seen in Fig~\ref{fig:norobustkernel} in the red ellipses. The emergency map is better fitted when using a Huber kernel, as in Fig~\ref{fig:Huber}, but still does not represent reality. 
When using only DCS, the SLAM map is corrupted since the error of some edges is scaled down without first converging toward an approximate solution (Fig~\ref{fig:dcs}). The rough scale of the emergency map introduces high errors in some link-edges and, with more than 50\% of wrong link-edges, DCS alone can not converge toward the optimal solution.  

\subsection{First eigenvalue of prior-edges' covariance.}
\label{subsec:priorcov}

\begin{figure}[t]
\centering
\begin{tabular}{cc}
        \subfloat[No optimization. The SLAM map is simply drawn over the prior.]{\includegraphics[height = 3.5cm]{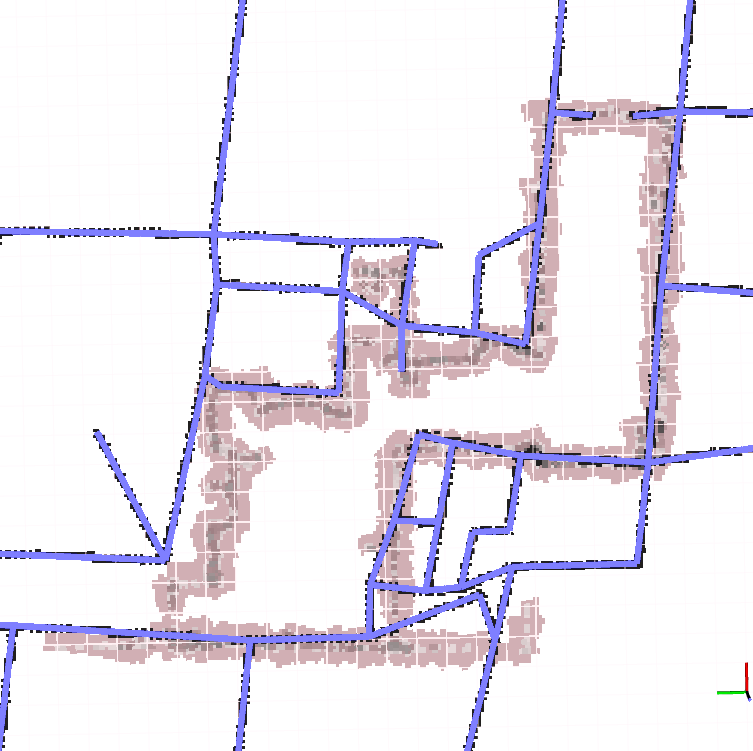}\label{fig:nopti}} &
        \subfloat[Using 1\% of a prior-edge length to calculate its covariance. Red circles: the prior is too rigid to fit the SLAM map.]{\includegraphics[height = 3.5cm]{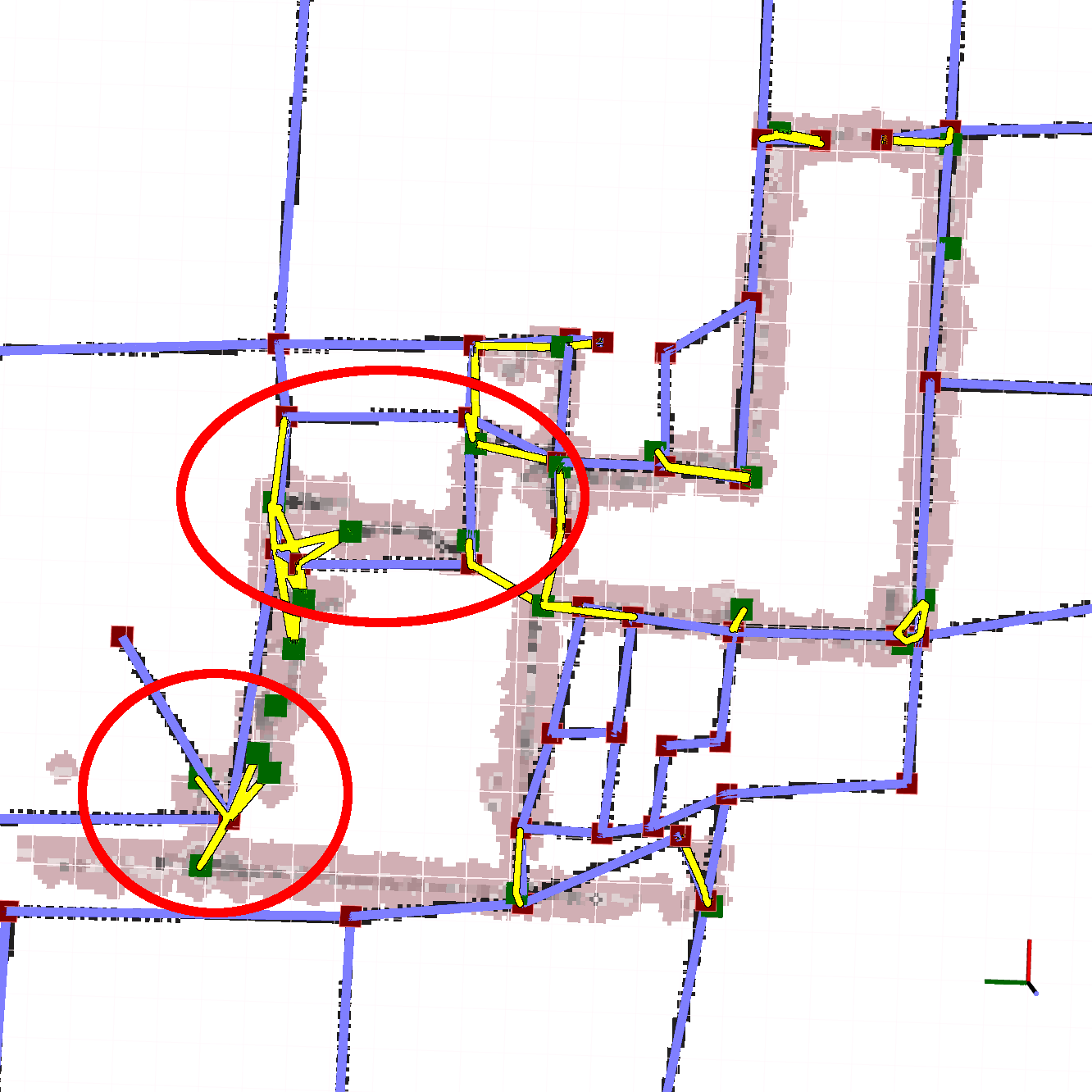}\label{fig:1}} \\
        \subfloat[Using 50\% of a prior-edge length to calculate its covariance. The prior and SLAM map are correctly fitted together.]{\includegraphics[height = 3.5cm]{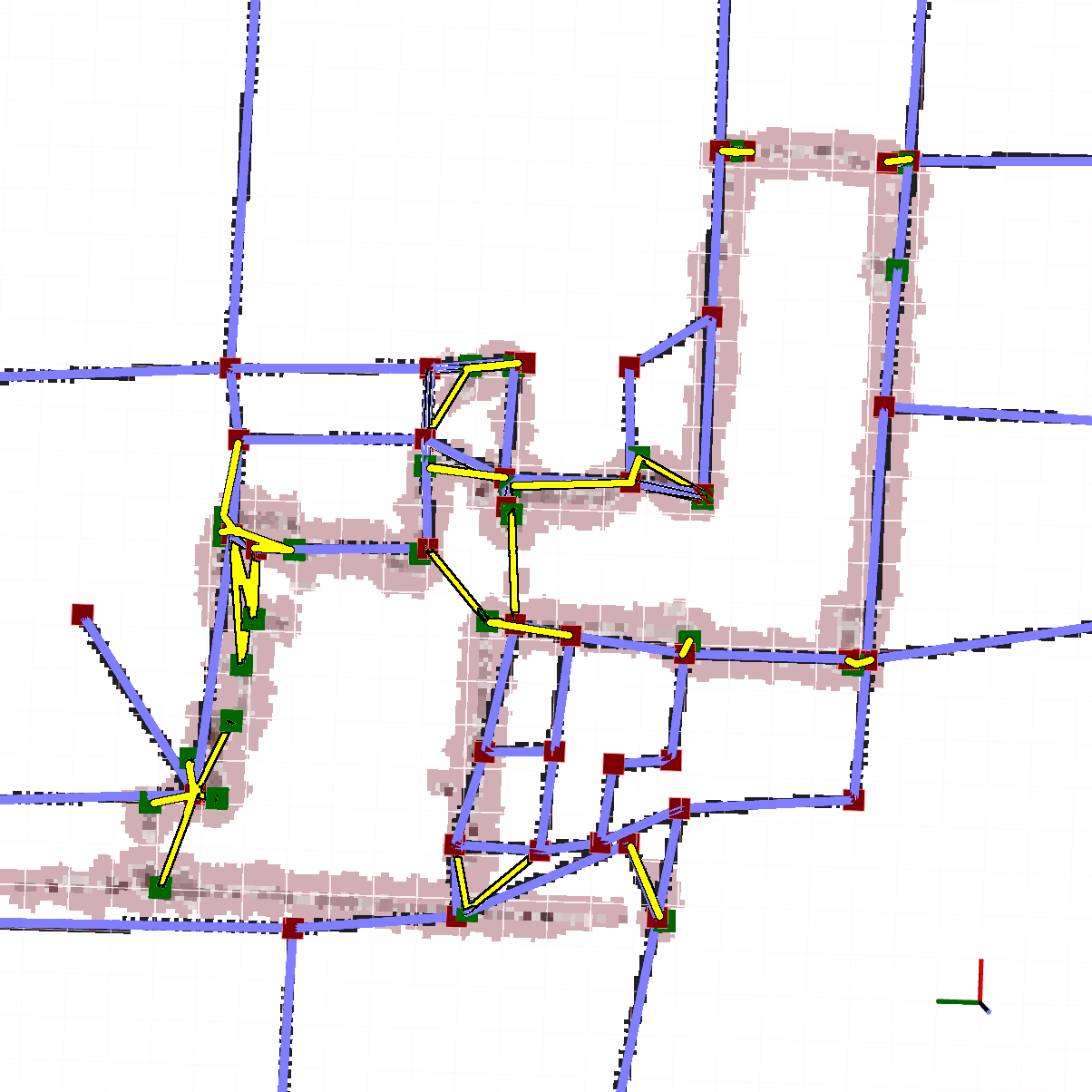}\label{fig:50}} &        
        \subfloat[Using 100\% of a prior-edge length to calculate its covariance. The prior was not rigid enough and is uncorrectly fitted on the SLAM map due to wrong link-edges.]{\includegraphics[height = 3.5cm]{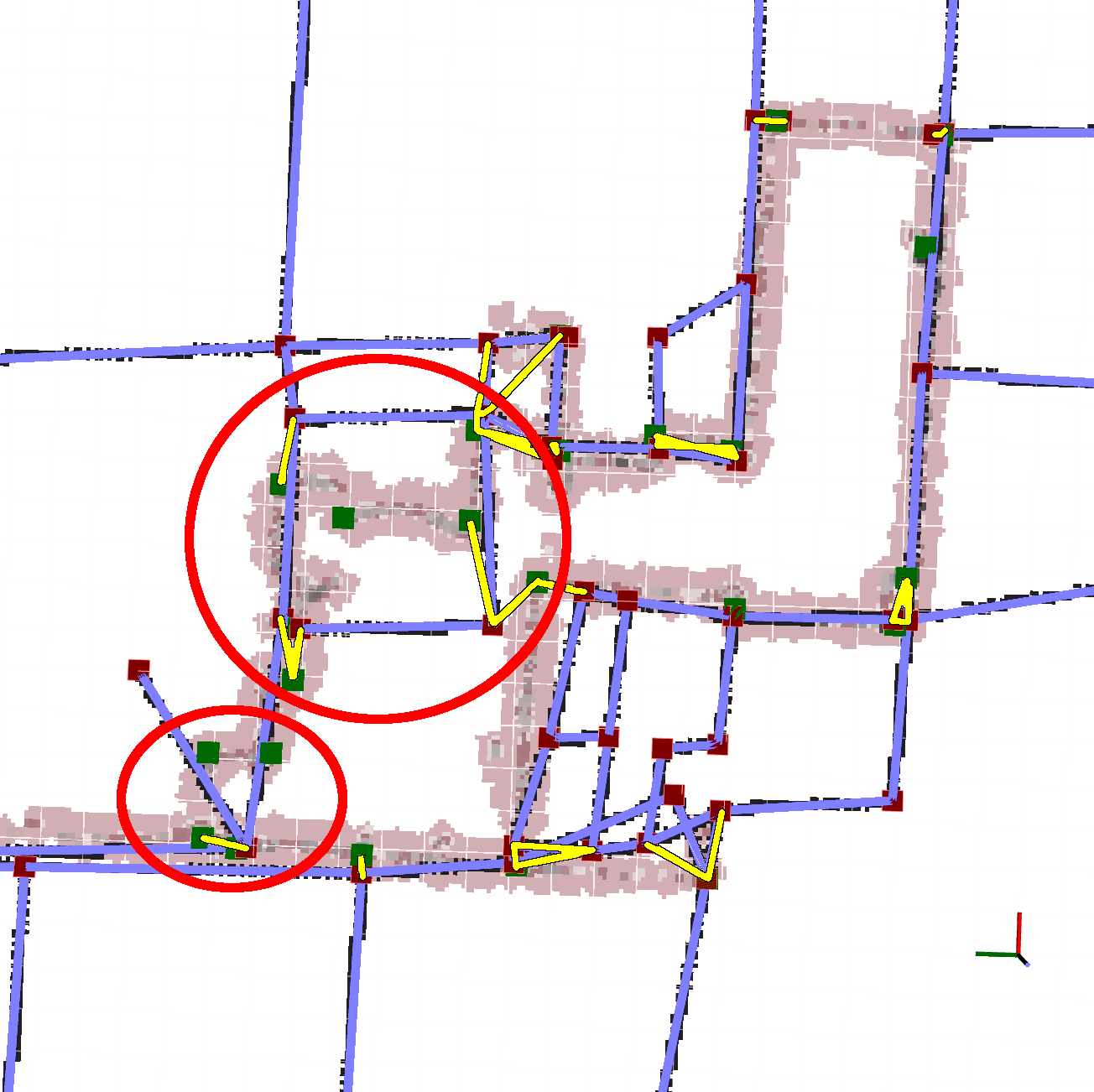}\label{fig:100}} 
\end{tabular}
\caption{Results of the optimization with different percentage of the egde's length used for the prior's covariance.}
\label{fig:priorcovtest}
\vspace{-6.5mm}
\end{figure}

We evaluate the influence of the eigenvalue corresponding to the eigenvector along each prior-edge (discussed in Section~\ref{subsubsec:elemerg}). This parameter controls how likely prior-edges are to extend or shrink.

As seen in Fig~\ref{fig:1}, using 1\% of the edge's length leads to the prior map not being changed but translated and rotated to fit the SLAM map, without correcting its inaccuracies. 
Using 100\% of the edge's length gives too much flexibility to the prior-edges and leads the ACG to a suboptimal solution, where walls from the emergency map don't match equivalent walls in the prior map, as can be seen in the red circles in Fig~\ref{fig:100}. 
Using 50\% gives the best result, since the prior is fitted onto the SLAM map, and incorrect corner correspondences do not influence the result, as seen in Fig~\ref{fig:50}.

\subsection{Maximum number of outlier link-edges}
\label{subsec:outliers}

\begin{figure}[t]
\vspace{+2mm}
    \begin{center} 
      \begin{tikzpicture}[gnuplot]
\tikzset{every node/.append style={font={\footnotesize}}}
\gpcolor{rgb color={0.000,0.620,0.451}}
\gpsetlinetype{gp lt border}
\gpsetdashtype{gp dt solid}
\gpsetlinewidth{1.00}
\draw[gp path] (0.460,0.985)--(0.460,1.165);
\node[gp node center] at (0.460,0.677) {$60$};
\draw[gp path] (1.396,0.985)--(1.396,1.165);
\node[gp node center] at (1.396,0.677) {$65$};
\draw[gp path] (2.332,0.985)--(2.332,1.165);
\node[gp node center] at (2.332,0.677) {$70$};
\draw[gp path] (3.267,0.985)--(3.267,1.165);
\node[gp node center] at (3.267,0.677) {$75$};
\draw[gp path] (4.203,0.985)--(4.203,1.165);
\node[gp node center] at (4.203,0.677) {$80$};
\draw[gp path] (5.139,0.985)--(5.139,1.165);
\node[gp node center] at (5.139,0.677) {$85$};
\draw[gp path] (6.075,0.985)--(6.075,1.165);
\node[gp node center] at (6.075,0.677) {$90$};
\draw[gp path] (7.010,0.985)--(7.010,1.165);
\node[gp node center] at (7.010,0.677) {$95$};
\draw[gp path] (7.946,0.985)--(7.946,1.165);
\node[gp node center] at (7.946,0.677) {$100$};
\draw[gp path] (0.460,3.661)--(0.460,0.985)--(7.946,0.985);
\gpcolor{rgb color={0.000,0.376,0.678}}
\gpsetdashtype{dash pattern=on 2.00*\gpdashlength off 5.00*\gpdashlength }
\gpsetlinewidth{2.00}
\draw[gp path](2.893,0.985)--(2.893,3.661);
\gpcolor{rgb color={0.867,0.094,0.122}}
\draw[gp path](2.332,0.985)--(2.332,3.661);
\gpcolor{color=gp lt color border}
\node[gp node center] at (4.203,0.215) {percentage of outlier link-edges};
\node[gp node right] at (6.478,3.327) {Success};
\gpcolor{rgb color={0.000,0.376,0.678}}
\gpsetdashtype{gp dt solid}
\draw[gp path] (6.662,3.327)--(7.578,3.327);
\draw[gp path] (1.770,2.769)--(3.080,2.769)--(3.267,2.769)--(2.144,2.769)--(1.396,2.769)%
  --(1.583,2.769)--(1.209,2.769)--(2.144,2.769)--(2.893,2.769)--(2.332,2.769)--(1.396,2.769)%
  --(1.957,2.769)--(0.741,2.769);
\gpsetpointsize{4.00}
\gppoint{gp mark 7}{(1.770,2.769)}
\gppoint{gp mark 7}{(3.080,2.769)}
\gppoint{gp mark 7}{(3.267,2.769)}
\gppoint{gp mark 7}{(2.144,2.769)}
\gppoint{gp mark 7}{(1.396,2.769)}
\gppoint{gp mark 7}{(1.396,2.769)}
\gppoint{gp mark 7}{(1.583,2.769)}
\gppoint{gp mark 7}{(1.209,2.769)}
\gppoint{gp mark 7}{(2.144,2.769)}
\gppoint{gp mark 7}{(2.893,2.769)}
\gppoint{gp mark 7}{(2.332,2.769)}
\gppoint{gp mark 7}{(1.396,2.769)}
\gppoint{gp mark 7}{(1.957,2.769)}
\gppoint{gp mark 7}{(0.741,2.769)}
\gppoint{gp mark 7}{(7.120,3.327)}
\gpcolor{color=gp lt color border}
\node[gp node right] at (6.478,3.019) {Failure};
\gpcolor{rgb color={0.867,0.094,0.122}}
\draw[gp path] (6.662,3.019)--(7.578,3.019);
\draw[gp path] (7.197,1.877)--(5.326,1.877)--(6.262,1.877)--(3.642,1.877)--(6.636,1.877)%
  --(5.326,1.877)--(4.390,1.877)--(2.519,1.877)--(6.823,1.877)--(3.080,1.877)--(6.636,1.877)%
  --(5.139,1.877)--(1.209,1.877);
\gppoint{gp mark 5}{(7.197,1.877)}
\gppoint{gp mark 5}{(5.326,1.877)}
\gppoint{gp mark 5}{(6.262,1.877)}
\gppoint{gp mark 5}{(6.262,1.877)}
\gppoint{gp mark 5}{(3.642,1.877)}
\gppoint{gp mark 5}{(6.636,1.877)}
\gppoint{gp mark 5}{(5.326,1.877)}
\gppoint{gp mark 5}{(4.390,1.877)}
\gppoint{gp mark 5}{(2.519,1.877)}
\gppoint{gp mark 5}{(6.823,1.877)}
\gppoint{gp mark 5}{(3.080,1.877)}
\gppoint{gp mark 5}{(6.636,1.877)}
\gppoint{gp mark 5}{(5.139,1.877)}
\gppoint{gp mark 5}{(1.209,1.877)}
\gppoint{gp mark 5}{(7.120,3.019)}
\gpcolor{rgb color={0.000,0.620,0.451}}
\gpsetlinewidth{1.00}
\draw[gp path] (0.460,3.661)--(0.460,0.985)--(7.946,0.985);
\gpdefrectangularnode{gp plot 1}{\pgfpoint{0.460cm}{0.985cm}}{\pgfpoint{7.946cm}{3.661cm}}
\end{tikzpicture}
      \vspace{-2mm}
      \caption{Success cases are in blue, and failure ones in red. The blue vertical bar is 1.5 standard deviation above the success mean, and the red one is 1.5 standard deviation under the failure mean.}
        \label{fig:normadist}
    \end{center}
    \vspace{-8mm}
\end{figure}
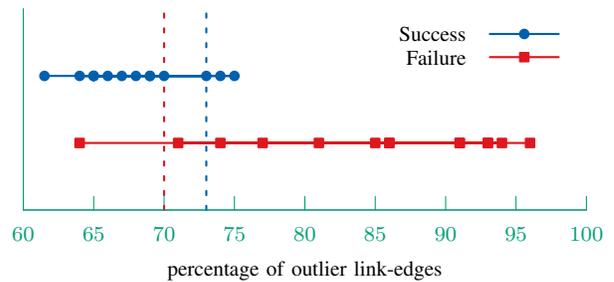

We estimated the maximum amount of wrong link-edges we can have in the graph before the optimization fails.
We ran 26 optimizations with added noise on the initial poses, totaling 13 successes and 13 failures, where a failure means that at least one prior landmark is not at the correct position in the SLAM map. The percentages of outliers for each optimization are presented in Fig~\ref{fig:normadist}.

To confirm that the number of outliers is correlated with the success of the optimization, we first test if the mean of the success sample is significantly less than the mean of the failures sample.
The z-score of the max and min of the percentage of outliers for success and failure samples are within a $3\sigma$ event and we can assume normality. We ran Welsch's t-test, which does not assume equality of variances. The sign of the t-statistic is important since we are testing a ``less than'' hypothesis of a one-tailed t-test. If $t < 0$ and $p_{value}/2 < 0.05$, we can reject the hypothesis that both distributions are similar. With our sample, $t = -4.756$ and the $p_{value} = 8.377 \cdot 10^{-5}$, showing a significant difference between the two distributions. 
For the failure case, 1.5 standard deviations under the mean represents $70\%$ of outliers, meaning $86\%$ of the failure cases have more than $70\%$ of outliers.
Thus, our method can usually handle up to $70\%$ outliers. Anecdotically, the optimization can fail under $70\%$, like one case in Fig~\ref{fig:normadist} but it should be noted that the final result only had 2 nodes over 13 wrongly matched and was very close to a success case. 
For the success case, 1.5 standard deviations above the mean is equivalent to $73\%$, which means that $86\%$ of the success cases have under $73\%$ of outliers. The optimization is likely to fail if there is more than $73\%$ of outliers, however, it can occasionally succeed.

\section{Real Navigation Scenario}
\label{sec:exp}

\begin{figure}[t]
\centering
\begin{tabular}{cc}
        \subfloat[The robot found a path to a room in the emergency map that was not yet visited by the robot.]{\includegraphics[height = 3.9cm]{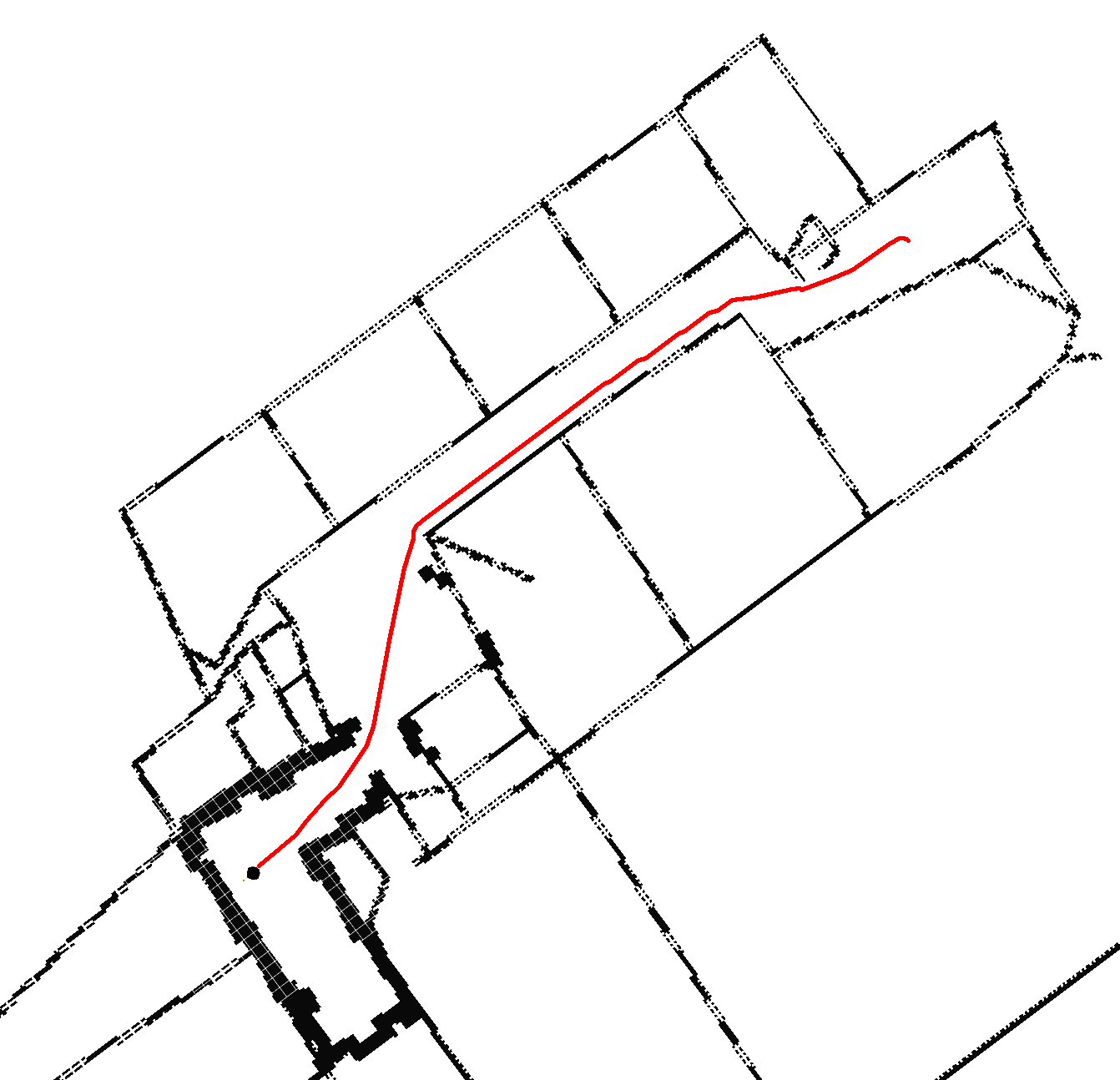}\label{fig:pathfinding_1node}} &
        \subfloat[The robot added a door that wasn't closed in the emergency map and the path can not be found anymore.]{\includegraphics[height = 3.9cm]{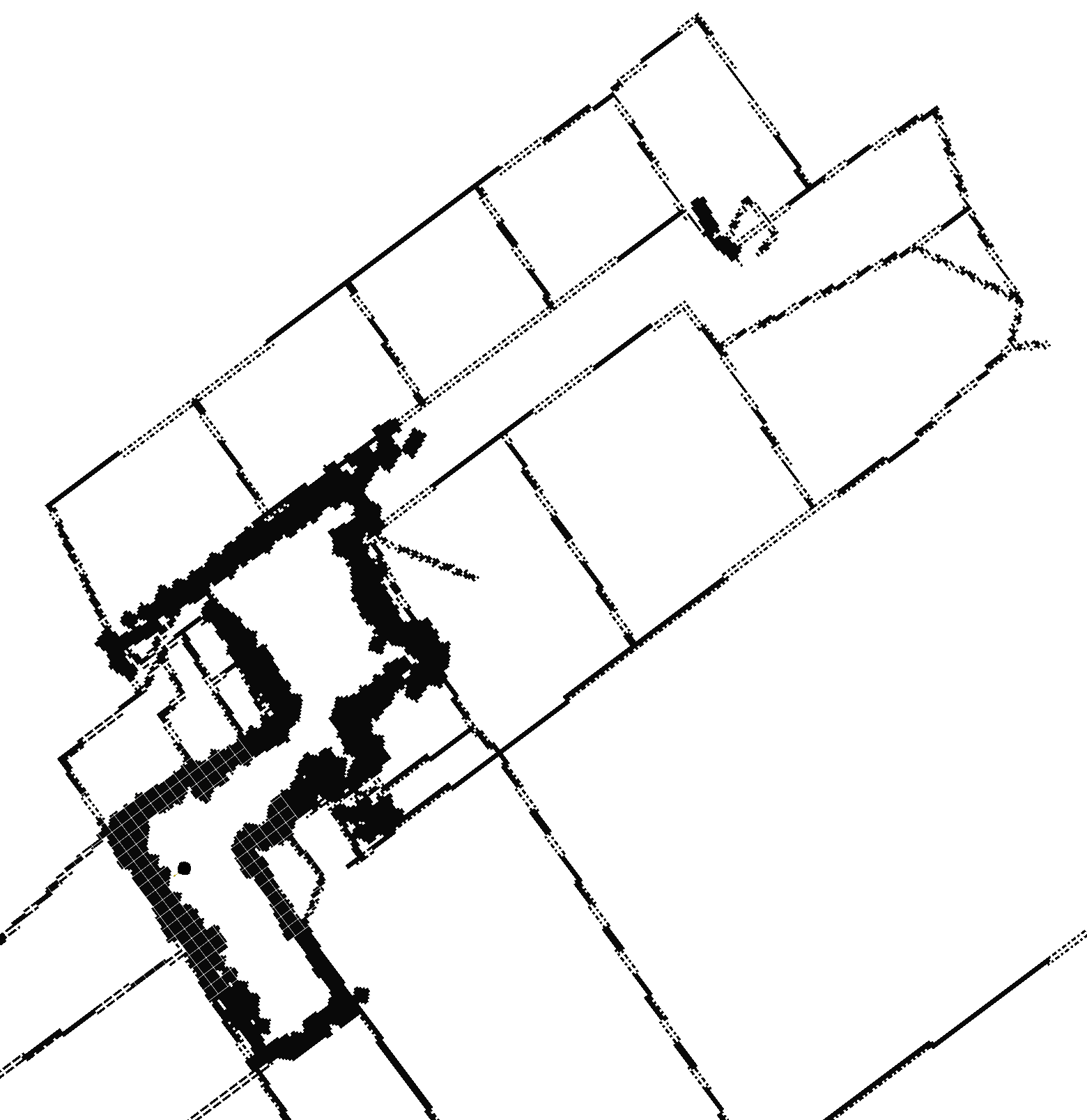}\label{fig:pathfinding_7nodes}}
\end{tabular}
\caption{Searching for a path at different exploration times.}
\vspace{-6mm}
\label{fig:pathsresults}
\end{figure}

We tested the ACG in a real navigation scenario. The test was conducted in the basement of Örebro University on the Taurob platform\footnote{\url{http://taurob.com}}, using a Velodyne and the same prior as before.
We used $2\,$m as the minimum distance between two corners before creating a link-edge, to get less than $71\%$ of outliers as seen in Section~\ref{subsec:outliers}, and 50\% as the eigenvalue for the prior-edges' covariances. The process runs in real-time on an Intel core i7 CPU at 2.30GHz, in about 1.3 seconds. Hence, the total time is slower than the time needed to create a partial NDT map.

To create an occupancy grid from the auto-complete graph, we fused all partial NDT maps positioned on their relative pose-node in one occupancy grid. Then, we draw the prior map by adding occupied cells where prior-edges are.
Using this map, it was possible to find paths to places the robot had not yet explored.
In Fig~\ref{fig:pathfinding_1node}, the robot has only been exploring for a short time but can find a path toward its destination, even though it has not explored that part of the environment yet. 
In Fig~\ref{fig:pathfinding_7nodes}, the robot has collected more information and now knows that the first path it found toward its goal is not practicable. 
A door is closed and the robot can not pass through it.

\section{Future work}

In the future, we plan on investigating improved correspondences between the emergency map and the SLAM map by introducing corner orientation. 
Also, we will evaluate our method using other SLAM datasets for which emergency maps, or other rough prior maps, are available.

\section{Conclusion}

We developed a formulation of graph-based SLAM, incorporating information from a rough prior,
which has uncertainties in scale and detail level. We also presented an optimization strategy adapted to this new graph formulation.

We use corners, in the prior and SLAM maps, as a common element to find correspondences, and create a graph fusing information from both modalities. To obtain robust optimization results, we first use a Huber kernel followed by DCS, allowing up to $70\%$ of wrong correspondences.

Contrary to other works~\cite{shah_qualitative_2013, persson_fusion_2008, kummerle_large_2011, parsley_towards_2010, parsley_exploiting_2011, vysotska_exploiting_2016}, we do not use the prior map to bind the SLAM map. We match the prior map onto the SLAM map to complete missing information and unexplored areas, while accounting for the uncertainties of the prior.
Experiments showed that the SLAM map completed with information from the emergency map enables the robot to navigate and plan the exploration while taking in account non-explored places.

\printbibliography %

\end{document}